\documentstyle[aclap]{article}
\author{Lena Str\"omb\"ack \\
        Department of Computer and Information Science \\
        Link\"oping University \\
        S-58185 Link\"oping, Sweden \\
        {\tt lestr@ida.liu.se}}

\title{User-Defined Nonmonotonicity in Unification-Based Formalisms}

\begin{document}
\maketitle
\bibliographystyle{acl}
\begin{abstract}
A common feature of recent unification-based grammar formalisms is
that they give the user the ability to define his own structures.
However, this possibility is mostly limited and does not include nonmonotonic
operations. In this paper we show how nonmonotonic operations can also be
user-defined by applying default logic \cite{reiter} and generalizing
previous results on nonmonotonic sorts \cite{youngrounds}.
\end{abstract}

\section{Background}

Most of the more recent unification-based formalisms, such as TFS
\cite{tfsref}, UD \cite{udref}, CUF \cite{cufref}, and FLUF \cite{coling94},
provide some possibility for the user to define constructions of his own.
This possibility can be more or less powerful in different formalisms.
There are, however, several constructions proposed as desirable
extensions
to unification grammars that cannot be defined in a general and
well-defined way in these
formalisms. One such class of constructions is those that have
some degree of nonmonotonic behaviour. Examples of such constructions are
any-values, default-values, and some constructions (e.g. constraining
equations, completeness and coherence) used in LFG \cite{lfgref}.

This paper describes a method that permits the user
to define such nonmonotonic constructions. This is done through
generalizing the work on nonmonotonic sorts
\cite{youngrounds}. This generalization results in
a default logic similar to
\cite{reiter}, but where subsumption and unification are used instead
of logical truth and consistency.

There are three main advantages to Young and Rounds'
work compared with other approaches to default unification
\cite{bouma90,bouma92,russel} which justify choosing it as a starting
point for this work. The first is the separation of definite and default
information, where Young and Rounds are more distinct than the other. The
second is that the nonmonotonic unification operation used is order
independent. This is achieved by separating the unification operation
from computing the nonmonotonic extension, which Young and Rounds call
explanation. This suggests that all the user needs to define when generalizing
the approach is how a sort is explained. Finally, there is a close
relationship to Reiter's (1980) default logic.

This paper starts by providing the minimal properties required of
a unification-based formalism when extending with nonmonotonic definitions.
I then describe the approach of user-defined nonmonotonicity
and
illustrate how some commonly used nonmonotonic constructions can be
defined within it. Finally I conclude with a discussion of
the relation to Reiter's default logic and computational properties
of the approach.

\section{Preliminaries}

There are two main properties that will be assumed of a
unification-based formalism in order to extend it with the
possibility of defining nonmonotonic constructions. The first, and most
important, is that we require a subsumption order on the set ${\cal S}$
of objects denoted by the formalism. Secondly it should be possible to define
inheritance hierarchies on the linguistic knowledge described by the formalism.

One very plausible subsumption order that can be used is the ordinary
subsumption lattice of feature structures. It is, however, possible to
use some other kind of subsumption order if that is more suitable for
the domain to be modelled by the formalism. Examples of other
subsumption orders that might be useful are typed feature
structures, feature structures extended with disjunction, or simply an
order based on sets and set inclusion.

In this paper the notation $a \sqsubseteq b$ is used
whenever $a$ subsumes $b$ (i.e. whenever $a$ ``is more specific than'' or
``contains more information than'' $b$). Consequently, $a \sqsubset b$
is used whenever $a \sqsubseteq b$ but $a \neq b$.

The subsumption order is assumed to be a semi-lattice and
permits computing a unifier, denoted $a \sqcap b$,
corresponding to the greatest lower bound, for every pair of
elements within it.
The element corresponding to the bottom of the order relation is
denoted {\it fail} and represents inconsistent information or unification
failure.

The second constraint placed on the formalism, the possibility of defining an
inheritance hierarchy, is not essential for the definition of
nonmonotonic operations. It is, however, very useful when
defining nonmonotonic constructions. The following notation will be used
for describing an inheritance hierarchy.

\begin{verse}

{\tt class} {\it the name of the class}{\tt ;} \\
{\tt isa} {\it its parent in the hierarchy}{\tt ;} \\
{\tt requires} {\it a structure}{\tt .}

\end{verse}

Thus, each member in the inheritance hierarchy is called a
class, which is defined by giving it a name and a parent
in the hierarchy. It is also possible to define some constraints,
called requirements, which
must hold for a class. These requirements can be both structures in the
subsumption order and nonmonotonic rules. The constraints
on classes are inherited through the hierarchy. Every
object in a class is assumed to contain at least the information given
by the constraints specified for it and all its ancestors. For simplicity
multiple inheritance between classes will not be allowed. This
means that two classes where none of them is a subclass of the other, will
always be considered inconsistent and thus yield a failure when unified.

\section{User-Defined Nonmonotonicity}

I will now describe how the work by Young and Rounds can be generalized
to allow the user to define nonmonotonic constructions.
The main idea in their approach is that each node in a
feature structure consists of a {\it nonmonotonic sort}. Such sorts
can contain two different kinds of information, the ordinary
monotonic information and a set of defaults.
If we assume that $\beta$ is defined as a default in
Young and Rounds' work then it is interpreted according to the rule:
if it is consistent to believe $\beta$ then believe $\beta$. In Reiter's
default logic this is expressed with the following normal default rule.

\[  \frac{:\beta}{\beta} \]

In this paper I want to allow the user to use other forms of
nonmonotonic inferences and not only the normal default rule given
above. Therefore, I will consider the general form of default rules.
An intuitive reading of a general default rule is, if $\alpha$ is
believed and it is consistent to believe $\beta$ then believe
$\gamma$. In default logic this is usually expressed as

\[  \frac{\alpha :\beta}{\gamma} \]

The next question is how such defined nonmonotonic rules are going to be
interpreted in a unification framework. In \cite{reiter}, a
rule like the one above could be applied
whenever $\alpha$ is true and $\beta$ is consistent with the information we
already have. If we assume that ${\cal V}$ represents the information already
given this means that the default rule can be applied whenever ${\cal
V}\sqsubseteq \alpha$ and ${\cal V} \sqcap \beta$ does not yield
unification failure. When the rule is applied the new information obtained
would be ${\cal V} \sqcap \gamma$.

In the approach described in this paper, the user is allowed to define
the actual nonmonotonic rule that
should be used for a particular operation by using the following
syntax.

\begin{verse}
\it

{\tt nonmon} name{\tt (}$parameter_1, \ldots parameter_n${\tt )}{\tt:}when
    $\alpha$ {\tt:}$\beta$ {\tt =>} $\gamma$

\end{verse}

In the syntax given above {\it name} assigns a name to the defined rule, and
thus allows the user to use nonmonotonic information when defining linguistic
knowledge. The parameters in the rule definition are variables, which
can be used within the actual default rule at the end
of the description. The user is assumed to assign the nonmonotonic
information contained in this rule to his linguistic knowledge by using
an expression of the form $name(parameter_1,\ldots parameter_n)$.

The {\it when} slot in the rule allows the user to decide when the
rule is going to be applied, or in Young and Rounds'
terminology, explained. I will make use of
two values for the when-slot, {\it immediate} and {\it posterior}. Immediate
means that the nonmonotonic rule is going to be applied each time
a full unification task has been solved or whenever
all information about an object in the defined inheritance hierarchy
has been retrieved. Posterior explanation means that
the explanation of the rule is postponed until reaching the result of
some external process, for example, a parser or generator. There is
however no hinder in excluding the use of other values here. One
could, for example, imagine cases where one would
want different nonmonotonic rules
to be explained after a completed parse, a generation, or after resolving
discourse referents.

Note that although the {\it when} slot in the definition of a
nonmonotonic rule allows the user to define when his rule is going to
be applied we will still have an order independent nonmonotonic
unification operator. This is the case because we follow Young and Rounds'
approach and separate the unification operation from the explanation
of a nonmonotonic rule. Therefore, what affects the final result
of a computation is when one chooses to explain default rules and
not the order of the unification operations occurring between such
explanations.

\section{Formal Definitions}

In this section I provide give the formal definitions for nonmonotonic
sorts and how nonmonotonic sorts are unified and explained. The
definitions are
generalizations of the definitions in Young and Rounds (1993).
The notation $a \sim b$ is used to denote the fact that
$a \sqcap b$ does not yield unification failure.

A nonmonotonic sort is a structure
containing both information from the basic subsumption order and
information about default rules to be explained at a later point in
the computation.

\begin{description}

\item[Definition 1]
A {\it nonmonotonic sort} is a pair $\langle s, \Delta \rangle$ where
$ s \in {\cal S} $ and $\Delta$ is a set of nonmonotonic rules of the
form $ \langle w, \alpha :\beta \Rightarrow \gamma \rangle$ where $w$ is
an atom and $\alpha$, $\beta$ and
$\gamma \in {\cal S}$. 	It is assumed that
for each nonmonotonic rule $\gamma \sqsubseteq \beta$,
$\alpha \sim s$, $\beta \sim s$, and $\gamma
\sqcap s \sqsubset s$.

\end{description}

As seen by the definition a nonmonotonic sort is considered to be a pair
of monotonic information from the subsumption order and nonmonotonic
information represented as a set of nonmonotonic rules. The user can
assign nonmonotonic information to a nonmonotonic sort by calling a
nonmonotonic definition as defined in the previous section. The actual
nonmonotonic rule occurring within the sort is a pair consisting
of the {\it when} slot and the last part of the
nonmonotonic definition, with the parameter variables instantiated
according to the call made by the user.

The second part of this definition contains some well-foundedness conditions
for a nonmonotonic sort. The
first condition ($\gamma \sqsubseteq \beta$) is a restriction similar
to the restriction to normal default rules in Reiter's
(1980) default logic. This restriction
ensures that the application of one default rule will never cause
previously applied default rules to be inapplicable. This makes the
procedure for application of defaults more efficient and
will be further discussed in section 6.

The next two conditions in the definition, $\alpha \sim s$ and $\beta
\sim s$, guarantee that the default rule is or can be applicable to
the nonmonotonic sort. The reason for only checking that $\alpha \sim
s$ instead of $s \sqsubseteq \alpha$
is that future unifications can restrict the value of $s$ into
something more specific than $\alpha$ and thus may make the default
rule applicable.

The last condition on a nonmonotonic sort, $\gamma \sqcap s \sqsubset
s$, may seem superfluous. The reason for including it is to ensure that
applying the default actually restricts the value of the sort.
Otherwise the default rule would have no effect and can be removed.
Note in particular that the above conditions on a nonmonotonic sort
implies that $\gamma$ may be {\it fail}.

Given the unification operation of objects within the subsumption order
and the definition of nonmonotonic sorts it is possible to define an
operation for nonmonotonic unification.

\begin{description}

\item[Definition 2]
The {\it nonmonotonic unification} ($\sqcap_N$) of two nonmonotonic sorts
$\langle s_1, \Delta_1 \rangle$ and $\langle s_2, \Delta_2 \rangle$ is
the sort $\langle s, \Delta \rangle$ where
\begin{itemize}
\item
$s=s_1\sqcap s_2$ and
\item
$\Delta = \{d \mid d=\langle w,\alpha :\beta \Rightarrow \gamma \rangle, \
d \in \Delta_1 \cup \Delta_2,\  \alpha \sim s,\  \beta \sim s,\  and \
\gamma \sqcap s \sqsubset s\}$
\end{itemize}
\end{description}

The nonmonotonic unification is computed by
computing the unification of the monotonic parts of the two sorts and
then taking the union of their nonmonotonic parts. The extra
conditions used when forming the union of the nonmonotonic parts of the
sorts are the same as in the definition of a nonmonotonic sort and
their purpose is to remove nonmonotonic rules that are no longer
applicable, or would have no effect when applied to the sort resulting
from the unification.

It is important to note that this generalization of the original
definition of nonmonotonic unification from Young and Rounds
(1993) preserves the property of order independence
for default unification.

When using nonmonotonic sorts containing nonmonotonic rules, we also need
to know how to merge the monotonic and nonmonotonic information within
the sort. I will use the terminology {\it w-application} for applying
one nonmonotonic rule to the sort and {\it w-explanation} when
applying all possible rules.

\begin{description}

\item[Definition 3]
The nonmonotonic rule  \\
 $\langle w, \alpha :\beta \Rightarrow \gamma
\rangle$ is {\it w-applicable} to $s \in \cal S$ if:
\begin{itemize}
\item
$s \sqsubseteq \alpha$
\item
$s \sim \beta$ or $s={\it fail}$
\item
$s \sqcap \gamma \sqsubset s$ or $s={\it fail}$
\end{itemize}
The result of the {\it w-application} is $\gamma \sqcap s$
\end{description}

Note that the $w$ in {\it w-application} should be considered as a variable.
This means that only nonmonotonic rules whose first component is $w$
are considered and that
it is possible to choose which nonmonotonic rules should be
applied in a particular point at some computation.

In addition note that the
restriction that $\gamma \sqsubseteq \beta$ in all nonmonotonic rules
and the special cases for $s={\it fail}$
ensures that the application of one nonmonotonic rule never destroys
the applicability of a previously applied rule. This reduces the
amount of work required when computing a w-explanation. Based on these
observations, a sufficient condition for w-explanation is defined as
follows.

\begin{description}

\item[Definition 4]
$t$ is a {\it w-explanation} of a nonmonotonic sort $\langle s,\Delta
\rangle$ if it can be computed in the following way:
\begin{enumerate}
\item
If $s={\it fail}$ or no $d \in \Delta$ is $w$-applicable then $t=s$ else
\item
Choose a $d=\langle w, \alpha : \beta \Rightarrow \gamma \rangle \in \Delta$
such that $d$ is $w$-applicable to $s$.
\item Let $s=s \sqcap \gamma$ and go to 1.
\end{enumerate}

\end{description}

As shown by the definition, a w-explanation is computed by choosing one
w-applicable default rule at a time and then applying it. Since the
definition of w-applicability and the condition that $\gamma
\sqsubseteq \beta$ in all nonmonotonic rules ensures that whenever a
nonmonotonic rule is applied it can never be inapplicable, there is no need
to check if the preconditions of earlier applied nonmonotonic rules
still hold.

Note also that the choice of which nonmonotonic rule to apply in each
step of a w-explanation is nondeterministic. Consequently, it is possible to
have conflicting defaults and multiple w-explanations for a
nonmonotonic sort.

Note also that the result of a
w-explanation is allowed to be {\it fail}. Another option would be to
interpret {\it fail} as if the
application of the nonmonotonic rule should not be allowed. However,
as seen in the next section, for many uses of nonmonotonic extensions
within unification-based formalisms, the aim is to derive failure if
the resulting structure does not fulfill some particular conditions.
This makes it important to allow {\it fail} to be the result of applying a
nonmonotonic rule.

\section{Examples}

In this section I will show how some of the most common nonmonotonic
extensions to unification-based grammar can be expressed by defining
rules as above. I will start with defining default values. This is done
by defining a nonmonotonic rule {\it default} for the class value, which
is assumed to be the most general class in a defined hierarchy. The rule
defined here is treated as the one in \cite{youngrounds}.

\begin{verse}
\tt
 class value; \\
 nonmon default(X):immediate
                         :X => X. \\
\end{verse}

This default rule can be used when defining verbs.
The rule is used for stating that verbs are active by
default. I also define the two Swedish verbs {\it skickade} ({\it sent})
and {\it skickades} ({\it was sent}) to illustrate how this rule works.

\begin{verse}
\tt
 class verb; \\
 isa value; \\
 requires [form: default(active)]. \\
\vspace{0.5 cm}
 class skickade; \\
 isa verb; \\
 requires [lex: skicka]. \\
\vspace{0.5 cm}
 class skickades; \\
 isa verb; \\
 requires [lex: skicka,
           form: passive]. \\
\end{verse}

While retrieving the information for these two verbs we will obtain
the following two feature structures containing nonmonotonic sorts:

\begin{verse}
For {\it skickade}: \\
{\tt
[lex: skicka,
 form: $\langle$[],$\lbrace \langle$immediate,
                                   :active
$\Rightarrow$ active $\rangle \rbrace \rangle$]} \\
\vspace{0.5 cm}
For {\it skickades}: \\
{\tt
[lex: skicka,
 form: $\langle$passive,$\lbrace \langle$immediate,
                                 :active
$\Rightarrow$ active $\rangle \rbrace \rangle$]} \\
\end{verse}

Since I have used {\it immediate} for the
when-slot in the definition of the default rule, this nonmonotonic rule will
be applied immediately after retrieving all information about a verb in the
hierarchy. For the two structures above, the default rule can be
applied for {\it skickade}, since {\it active} is consistent with [], but
not for {\it skickades}, since {\it active} and {\it passive} are
inconsistent. The result after applying {\tt immediate}-explanation to
the two structures above is shown below.

\begin{verse}
For {\it skickade}: \\
{\tt
[lex: skicka, form: active]} \\
\vspace{0.5 cm}
For {\it skickades}: \\
{\tt
[lex: skicka, form: passive]} \\
\end{verse}

Another nonmonotonic operation that has been used in LFG \cite{lfgref}
is the value constraint ({\it =c}) used to check whether a
substructure has a particular value after a completed parse.
The definition of value constraints as a nonmonotonic rule
makes use of negation, interpreted as negation as failure.

\begin{verse}
\tt
 class value;           \\
 nonmon =c(X):posterior
                  :$\lnot$X => fail.
\end{verse}

One use of value constraints in LFG is to
assert a condition that some grammar rules can only be used for passive
sentences. I will here assume that a representation for verbs
where passive verbs have the value {\it passive} for the attribute
{\it form}, but where other verbs have no value for this attribute.
In the syntax used in this paper the constraint that a particular
grammar rule can only
be used for passive verbs would be expressed as below:

\begin{verse}
\tt
[form: =c(passive)]
\end{verse}

This would result in the nonmonotonic sort:

\begin{verse}
{\tt
[form: $\langle$[],$\lbrace \langle$posterior,
                 :$\lnot$passive
$\Rightarrow$ fail $\rangle \rbrace \rangle$]}
\end{verse}

As seen by the definition of {\it =c}, the explanation for this
nonmonotonic sort is postponed and is assumed to be computed
after finding a parse for some sentence. This implies
that the only case where this rule would not apply, and thus not give
{\it fail} as a result, is when the value of {\it form} actually is {\it
passive}. For all other values of form, we would have something that
is consistent with $\lnot \it passive$ and thus the nonmonotonic rule
will derive failure when applied.

The next nonmonotonic structure I want to discuss is {\it any-values}.
The inheritance hierarchy is used to be able to define any-values
in a simple way.

\begin{verse}
\tt
 class value. \\
\vspace{0.5 cm}
 class none; \\
 isa value.  \\
\vspace{0.5 cm}
 class any\_value;  \\
 isa value.  \\
 nonmon any():posterior :any\_no\_value => fail. \\
\vspace{0.5 cm}
 class any\_no\_value; \\
 isa any\_value. \\
\end{verse}

In this small hierarchy it is assumed that all possible values of a
structure is a subtype of {\it value}. We then divide this into
{\it none}, which represents that a structure cannot have any value
and {\it any\_value} which contains all actual values. The class {\it
any\_value} is then further divided into a class called {\it any\_no\_value},
which only contains this single value, and the actual values of a
structure. The class {\it any\_no\_value} should not be used when defining
linguistic knowledge. However,
when applying the default rule a value that has not been instantiated
is compatible with this {\it any\_no\_value}. Therefore the default rule can
make the conclusion that the structure is inconsistent, which is what
we desire. Note that, as soon as a value has been
further instantiated into a 'real' value, it will no longer be consistent
with {\it any\_no\_value}, and the nonmonotonic rule cannot apply.
Two examples will further illustrate this.

\begin{verse}
The nonmonotonic sort: \\
{\tt $\langle$[],
$\lbrace \langle$ posterior, :any\_no\_value
$\Rightarrow$ fail $ \rangle \rbrace \rangle$} \\
will be {\it posterior}-explained to: \\
{\tt fail} \\
\vspace{0.5 cm}
While the sort: \\
{\tt $\langle$[lex: kalle],
$\lbrace \langle$ posterior, :any\_no\_value
$\Rightarrow$ fail $ \rangle \rbrace \rangle$} \\
will be {\it posterior}-explained to: \\
{\tt [lex: kalle]}
\end{verse}

The last nonmonotonic operations I want to discuss are completeness
and coherence as used in LFG. To be able to define these operations I
assume the inheritance hierarchy above, without the
nonmonotonic definition of {\it any}. I will, instead, make use of the
two nonmonotonic definitions below.

\begin{verse}
\tt
 class value;  \\
 nonmon coherence(A):immediate :[A: none] => [A: none]; \\
 nonmon completeness(A):posterior :[A: any\_no\_value] => fail. \\
\end{verse}

The first of these rules is used to check coherence, and the effect
is to add the value {\it none} to each attribute that has been
defined to be relevant for coherence check, but has not been assigned
a value in the lexicon. The second rule is used for checking
completeness and it works similarly to the any-definition
above.

Finally, I will show how a fragment of a lexicon can be defined
according to these rules. Note that in the definition of the
transitive verb, the value {\it any\_value} is given to the appropriate
attributes. This means that they are inconsistent with {\it none}, and thus,
the coherence rule cannot be applied.

\begin{verse}
\tt
 concept verb; \\
 isa any\_value; \\
 requires coherence(subj) $\wedge$ coherence(obj) $\wedge$ ...; \\
 requires completeness(subj) $\wedge$ completeness(obj) $\wedge$ .... \\
\vspace{0.5 cm}
 concept transitiveverb; \\
 isa verb; \\
 requires [subj: any\_value, obj: any\_value]. \\
\end{verse}

\section{Relation to Default Logic}

In this section I will discuss the relation of this work to
Reiter's (1980) default logic.
There will also be some discussion on the computational
properties and limitations of the given approach.

Compared with Reiter's default logic, our notion of nonmonotonic
sorts corresponds to default theories. Unification of nonmonotonic
sorts would then correspond to merging two default theories into one
single theory and our notion of explaining a nonmonotonic sort corresponds
to computing the extension of a default theory in default logic.

In default logic there is often a restriction to normal-default
theories since non-normal default theories are not even
semi-decidable. The restriction in our nonmonotonic rules
that $\gamma \sqsubseteq \beta$ is similar to the restriction into
normal default rules and captures the important property, that
the application of one nonmonotonic rule should not affect the
applicability of previously applied rules.
The decidability of the nonmonotonic rules defined here is, however, highly
dependant on the given subsumption order. In particular it is dependent on
having a decidable unification operation and subsumption check.

As mentioned previously there is with nonmonotonic sorts, as well as
normal default logic, a possibility of conflicting
defaults and thus multiple nonmonotonic extensions for a structure.
One difference is that nonmonotonic sorts allow that the
application of a nonmonotonic rule
leads to {\it fail}, i.e. an inconsistent structure,
while default logic does not allow this outcome. However, since
{\it fail} is allowed as a valid explanation for a nonmonotonic sort,
there is, as for normal default logic, always at least one
explanation for a sort.

The two following examples will illustrate
the difference between nonmonotonic rules giving multiple extensions
and nonmonotonic rules giving a single explanation {\it fail}.

\begin{description}
\item{Example a}

\[ \frac{{\rm :}[a{\rm :}1]}{[a{\rm :}1 \  b{\rm :}1]} \hspace{1.5 cm}
   \frac{{\rm :}[c{\rm :}2]}{[b{\rm :}2 \  c{\rm :}2]} \]

\item{Example b}

\[ \frac{{\rm :}[a{\rm :}1]}{[a{\rm :}1 \  b{\rm :}1]} \hspace{1.5 cm}
   \frac{{\rm :}[b{\rm :}2]}{[a{\rm :}2 \  b{\rm :}2]} \]

\end{description}

In example $a$ the application of one rule, does not make the
other inapplicable. Thus the only explanation for a structure is achieved by
applying both these two rules and results in {\it fail}. In example $b$,
however, the
application of one of the rules would block the application of the
other. Thus, in this case there are two explanations for the structure
dependant on which of the rules that has been applied first. Note that even
though there is an order dependency on
the application order of nonmonotonic rules
this does not affect the order independency
on nonmonotonic unification between
application of nonmonotonic rules.

Allowing multiple extensions gives a higher computational complexity
than allowing only theories with one extension. Since it
is the user who defines the actual nonmonotonic theory multiple
extensions must be allowed and it must be considered a task
for the user to define his theory in the way he prefers.

\section{Improvements of the Approach}

I will start with two observations regarding the definitions given in
section 3. First, it is possible
to generalize these definitions to allow the first
component of a nonmonotonic sort to contain substructures
that are also nonmonotonic sorts. With the generalized versions of the
definitions explanations that simultaneously
explain all substructures of a nonmonotonic sort will be considered.
Note that the explanation of default
rules at one substructure might affect the explanation of rules at
other substructures. Therefore the order on which nonmonotonic rules
at different substructures are applied is important and all possible
application orders must be considered.

Considering unification of nonmonotonic sorts
it is not necessary to simplify the nonmonotonic part of the
resulting sort. $\Delta = \Delta_1 \cup \Delta_2$ can be defined as an
alternative to the given definition. This alternate definition is
useful for applications where the simplification of
nonmonotonic sorts by each unification is expected
to be more expensive than the
extra work needed to explain a sort whose nonmonotonic part is not
simplified.

As stated previously, nonmonotonic sorts allow multiple explanations of a
nonmonotonic sort.
If desired, it would be fairly easy to add priorities to the
nonmonotonic rules, and thus induce a preference order on explanations.

One further example will illustrate that it is also possible to define
{\it negation as failure} with nonmonotonic rules.
An intuitive interpretation of the defined rule below is
that if $X$ is believed (${\cal V}\sqsubseteq X$), failure should be
derived.

\begin{verse}
\tt
 nonmon not(X):immediate X => fail;
\end{verse}

However, if this definition is to be really useful we must also allow
one definition of a nonmonotonic rule to make use of other
nonmonotonic rules. In our original definition we said that the nonmonotonic
rule above should be applied if ${\cal V}\sim \lnot X$. This can be
generalized to the case where $\lnot X$ is a nonmonotonic rule if we extend
the definition of $\sim$ to also mean that the application (or explanation)
of the $not$ rule at this node does not yield failure. However, this
generalization is outside default logic. Therefore, its
computational properties are unclear and needs more investigation.

\section{Conclusion}

In this paper I have proposed a method allowing the user to define nonmonotonic
operations in a unification-based grammar formalism. This was done by
generalizing the work on nonmonotonic sorts \cite{youngrounds} to allow not
only normal defaults rules but general default rules that are defined by the
user. The method has a very close relation to Reiter (1980). We
also noted that the method can be applied to all domains of structures
where we have a defined subsumption order and unification operation.

The generality of the approach was demonstrated by defining some of the
most commonly used nonmonotonic operations. We also gave formal
definitions for the approach and provided a discussion on its computational
properties.

\section*{Acknowledgments}

This work has been supported by the Swedish Research Council for
Engineering Sciences (TFR). I would also like to thank
Lars Ahrenberg and Patrick Doherty for comments on this work and
Mark A. Young for providing me with much-needed information
about his and Bill Rounds' work.

\end{document}